\documentclass[11pt,a4paper]{article}
\usepackage[hyperref]{naaclhlt2018}
\usepackage{times}
\usepackage{latexsym}
\usepackage{amsmath,graphicx}
\usepackage{url}
\usepackage{multirow,bm}

\DeclareMathOperator{\GRU}{GRU}

\aclfinalcopy 

\title{Why Do Neural Dialog Systems Generate Short and Meaningless Replies?\\
	A Comparison between Dialog and Translation}

\author{Bolin Wei,\thanks{\ \ Equal contribution.}\,\,$^\dag$ Shuai Lu,$^{*\dag}$ Lili Mou,$^\ddag$ Hao Zhou,$^\P$ Pascal Poupart,$^\ddag$ Ge Li,$^\dag$ Zhi Jin$^\dag$\\
	$^\dag$Key Laboratory of High Confidence Software Technologies (Peking University)\\
	Ministry of Education, China; Software Institute, Peking University, China\\
	$^\ddag$David R. Cheriton School of Computer Science, University of Waterloo\quad
	$^\P$Toutiao AI Lab\\
{\em \{bolin.wbl, doublepower.mou\}@gmail.com\quad \{shuai.l, lige, zhijin\}@pku.edu.cn}\\
{\em zhouhao.nlp@bytedance.com\quad ppoupart@uwaterloo.ca}
}
\date{}

\begin{document}
\maketitle
\begin{abstract}
  This paper addresses the question: Why do neural dialog systems generate short and meaningless replies?
  We conjecture that, in a dialog system, an utterance may have multiple equally plausible replies, causing the
  deficiency of neural networks in the dialog application. We propose a systematic way to mimic the dialog scenario
  in a machine translation system, and manage to reproduce the phenomenon of generating short and less meaningful sentences in the translation setting, showing evidence of our conjecture.
\end{abstract}

\section{Introduction}

Open-domain human-computer dialog systems are attracting increasing attention in the NLP community. With the development of deep learning, sequence-to-sequence (Seq2Seq) neural networks, or more generally encoder-decoder frameworks, are among the most popular models for utterance generation in dialog systems~\cite{NRM,diversity,seq2BF,variational}.

Historically, Seq2Seq-like models are first designed for machine translation~\cite{seq2seq,attention} and later widely applied to image captioning~\cite{caption}, text summarization~\cite{summarization}, etc. When adapted to open-domain dialog systems, however, Seq2Seq models are less satisfactory. A severe problem is that the Seq2Seq model tends to generate short and meaningless replies, e.g.,  ``I don't know''~\cite{diversity} and ``Me too''~\cite{seq2BF}. They are universally relevant to most utterances, called \textit{universal replies} in \newcite{seq2BF}, and hence less desired in real-world conversation systems.

In previous studies,  researchers have proposed a variety of approaches to address the problem of universal replies, ranging from heuristically modified training objectives~\cite{diversity}, diversified decoding algorithms~\cite{DBS}, to content-introducing approaches~\cite{seq2BF,topic}. 

Although universal replies have been alleviated to some extent, there lacks an empirical explanation to the curious question: \textit{Why does the same Seq2Seq model tend to generate shorter and less meaningful sentences in a dialog system than in a machine translation system?}

Considering the difference between dialog and translation data, our intuition is that the dialog system encounters a severe unaligned problem: an utterance may have multiple equally plausible replies, which may have different meanings. On the contrary, the translation datasets typically have a precise semantic matching between the source and target sides. This conjecture is casually expressed in our previous work~\cite{seq2BF}, but is not supported by experiments.

In this paper, we propose a method to verify the conjecture by mimicking the unaligned scenario in machine translation datasets. We propose to shuffle the source and target sides of the translation pairs to artificially build a conditional distribution of target sentences with multiple plausible data points. By doing so, we manage to shorten the length and lower the ``information'' of generated sentences in a Seq2Seq machine translation system. This shows evidence that the unaligned problem could be one reason that causes short and meaningless replies in neural dialog systems.

To summarize, this paper systematically compares Seq2Seq dialog and translation systems, and provides an explanation to the question: Why do neural dialog systems tend to generate short and meaningless replies? Our study also sheds light on the future development of neural dialog systems as well as the application scenarios where Seq2Seq models are appropriate.

In the rest of this paper, we first describe our conjecture in Section~\ref{sec:conjecture}. Then we design the experimental protocol in Section~\ref{sec:protocol} and present results in Section~\ref{sec:results}. Finally, we conclude with discussion in Section~\ref{sec:conclusion}.

\section{Conjecture}\label{sec:conjecture}

We hypothesize that \textit{given a source sequence, the conditional distribution of the target sequence having multiple plausible points is one cause of the deficiency of Seq2Seq models in dialog systems}.

Let us denote the source sequence by $\bm s=s_1, s_2, \cdots, s_{|\bm s|}$ and the target sequence by $\bm t=t_1,t_2,\cdots, t_{|\bm t|}$. Both (orthodox) training and prediction objectives are to maximize $p_{\bm \theta}(\bm t|\bm s)$, where the conditional probability $p_{\bm \theta}(\cdot|\cdot)$ is modeled by a Seq2Seq neural network with parameters $\bm \theta$.

\begin{figure}[!t]
	\centering
	\resizebox{.8\linewidth}{!}{
	\includegraphics[width=\linewidth]{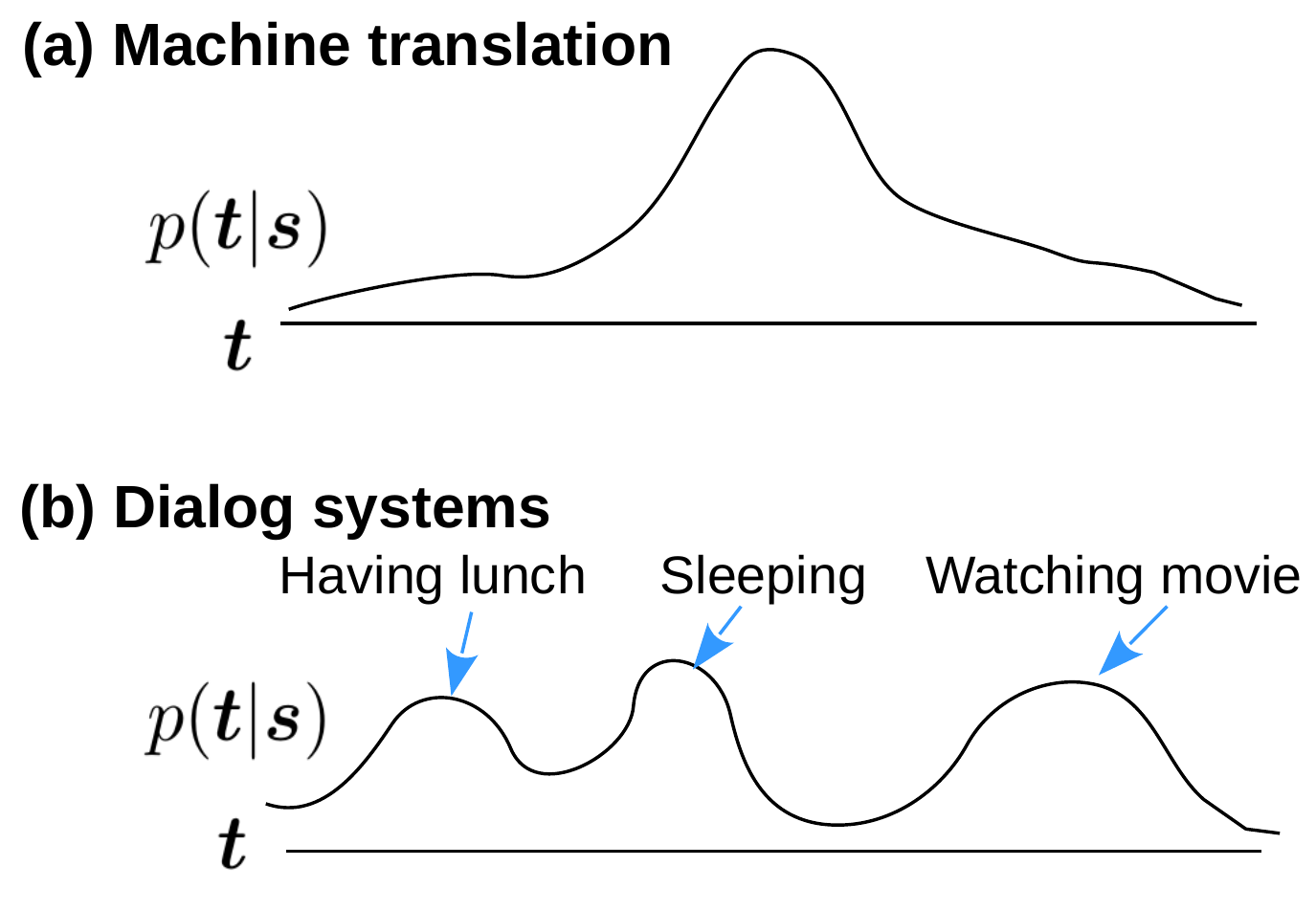}
	}\vspace{-.4cm}
	\caption{The conditional distribution $p(\bm t|\bm s)$ in (a) machine translation and (b) dialog systems, where we consider an analog of continuous random variables. More rigorously speaking, $p(\bm t|\bm s)$ is peaked at one or a few similar sentence(s) in machine translation because source and target information generally aligns, whereas an utterance can have multiple plausible replies in dialog systems.}\label{fig:distribution}\vspace{-.2cm}
\end{figure}

In a machine translation system, the source and target information generally aligns well, although some meanings could have different expressions. Figure~\ref{fig:distribution}a shows a continuous analog of $p(\bm t|\bm s)$.

In an open-domain dialog system, however, an utterance can have a variety of replies that are (nearly) equally plausible.
For example, given a user-issued utterance ``What are you going to do?'' there could be multiple replies like ``having lunch,'' ``watching movies,'' and ``sleeping,'' shown in Figure~\ref{fig:distribution}b with an analog of continuous random variables.
There is no particular reason why one reply should be favored over another without further context. Even with context, this problem could not be fully solved because of the true randomness of dialog. 

\begin{table*}
	\centering
	\small
	\begin{tabular}{|c|l||r|r|r|r|r|}
		\hline
		\multicolumn{2}{|c||}{Setting}  & BLEU & BLEU-1 & BLEU-2 & BLEU-3 & BLEU-4 \\
		\hline\hline
		Dialog  & Seq2Seq   &   1.84& 15.1& 2.40& 1.02& 0.66   \\
		\hline\hline
		\multirow{5}{*}{Translation} 
		& Seq2Seq          & 27.2 & 60.2 & 33.4 & 20.9 & 13.6 \\
		& +shuffle 25\%    & 24.4 & 56.2&30.3&18.8&12.0\\
		& +shuffle 50\%    & 21.1& 52.8&26.8&16.0&10.0\\
		& +shuffle 75\%    &17.2& 48.2&23.2&13.4&8.10\\
		& +shuffle 100\%   &.024 & 12.5&.189&0.00&0.00\\
		\hline
	\end{tabular}\vspace{-.3cm}
	\caption{BLEU scores of dialog and translation systems.}\label{tab:bleu}
	\vspace{-.2cm}
\end{table*}

The above is, perhaps, the most salient difference between dialog and translation datasets. While it is tempting to think of Seq2Seq's performance in this way~\cite{seq2BF}, there does not exist a practical approach to verify the conjecture.

\section{Experimental Protocol}\label{sec:protocol}

\subsection{Mimicking a ``Dialog Scenario'' in the Machine Translation}
We propose to mimic the ``unaligned'' property in a translation dataset by shuffling the source and target pairs. 
This ensures the resulting conditional distribution $p(\bm t|\bm s)$ to have multiple plausible data points, whereas other settings of translation remain unchanged, making a rigorous controlled experiment.

Formally speaking, let $\{(\bm s^{(n)}, \bm t^{(n)})\}_{n=1}^N$ be the training dataset in a translation setting, where $(\bm s^{(n)}, \bm t^{(n)})$ is a particular data point containing a source and target sentence pair; in total we have $N$ data points.

The shuffled dataset is $\{(\bm s^{(n)}, \widetilde{\bm t}^{(n)})\}_{n=1}^N$, where $\widetilde{\bm t}^{(n)}={\bm t}^{(\tau(n))}$
and $\tau(1), \cdots, \tau(N)$ is a random permutation of $1, 2, \cdots, N$. In this way, we artificially construct a conditional target distribution $p(\widetilde{\bm t}^{(n)}|\bm s^{(n)})$ that allows multiple plausible sentences conditioned on a particular source sentence. 

Notice that, for the sake of constructing a distribution where the target sentences can have multiple plausible data points, there is no need to generate multiple random target sentences for a particular source sentence. In fact, it is preferred NOT, so that the experiment is more controlled. In the case where we generate a single target sentence $\widetilde{\bm t}^{(n)}={\bm t}^{(\tau(n))}$ for a source sentence $\bm s^{(n)}$, $\{\widetilde{\bm t}^{(n)}|\bm s^{(n)}\}_{n=1}^N$ can still be viewed as samples from the marginal (unconditioned) distribution $p(\bm t)$, and thus the desired ``unaligned'' property is in place.

It is straightforward to shuffle a subset of the translation dataset. Details are not repeated here. This helps to analyze how Seq2Seq models behave when the ``unaligned'' problem becomes more severe. 

It should also be mentioned that the shuffling trick is previously used in \newcite{shuffle} to compare the robustness of Seq2Seq models and phrase-based statistical machine translation in terms of BLEU scores. Our paper contains a novel insight that shuffling datasets mimics the unaligned property in dialog datasets, which facilitates the comparison between Seq2Seq dialog and translation systems.

\subsection{The Seq2Seq Model and Datasets}

We adopted a modern Seq2Seq model (with an attention mechanism) as the neural network for both dialog and translation systems.
The encoder is a bidirectional recurrent neural network with gated recurrent units (GRUs), whereas the decoder comprises two GRU state transition blocks and an attention mechanism in between~\cite{toolkit}.\footnote{Code downloaded from \url{https://github.com/EdinburghNLP/nematus}}

For the dialog system, we used the Cornell Movie-Dialogs Corpus dataset,\footnote{Available at \url{https://www.cs.cornell.edu/~cristian/Cornell_Movie-Dialogs_Corpus.html}} containing 221k samples. For machine translation, we used the WMT-2017 dataset\footnote{Available at \url{http://data.statmt.org/wmt17/translation-task/preprocessed/de-en/}} and focus on English-to-Germany translation; it contains 5.8M samples.

We first tried a normal machine translation setting and achieved comparable results to a baseline in \newcite{baseline}; thus our replication of the machine translation system is fair. In all settings, we used the same model and hyperparameters so that our comparison is also fair.

Appendix~\ref{appdx:setup} provides detailed model description and experimental setup.

\begin{table*}
	\centering
\small
	\begin{tabular}{|c|l||r|r||r|r|}
		\hline
		\multicolumn{2}{|c||}{\multirow{2}{*}{Setting}}  & \multicolumn{2}{c||}{Length}  & \multirow{2}{*}{Neg.~log-prob.} & \multirow{2}{*}{Entropy} \\\cline{3-4}
		\multicolumn{2}{|c||}{\ }   & \# words &  \% of Ref&  &\\
		\hline\hline
		\multirow{2}{*}{Dialog}      & References         & 14.40 & 100 & 8.79 & 8.91\\
		\cline{2-6}
		& Seq2Seq  &  11.70&81.3&8.08&7.92\\
		\hline\hline
		\multirow{6}{*}{Translation} & References   & 
		21.47& 100 & 11.4&10.2\\
		\cline{2-6}
		& Seq2Seq      & 21.24& 98.9 & 11.1 &9.98 \\
		& shuffle 25\%    & 21.02 & 97.9& 10.9&9.81\\
		& shuffle 50\%    & 20.73 & 96.6 & 10.8&9.66\\
		& shuffle 75\%    & 19.89&92.6 & 10.6& 9.39\\
		& shuffle 100\%   & 15.88&74.0& 9.34&4.46\\
		\hline
	\end{tabular}\vspace{-.3cm}
	\caption{Average length (along with the percentage of references), negative log-probability, and entropy of dialog and translation systems.}\label{tab:stat}
\end{table*}

\begin{table}
	\centering
	\vspace{-.3cm}
	\resizebox{\linewidth}{!}{
		\begin{tabular}{|c|l||r|r|}
			\hline
			\multicolumn{2}{|c||}{\multirow{2}{*}{Setting}}& \multicolumn{2}{c|}{$R^2$ Correlation}\\
			\cline{3-4} \multicolumn{2}{|c||}{\ } & Encoder& Decoder \\
			\hline\hline
			Dialog  & Seq2Seq   &   .5095 & .1706  \\
			\hline\hline
			\multirow{5}{*}{Translation} 
			& Seq2Seq          & .9673 & .8734\\
			& +shuffle 25\%    & .9257 & .7241\\
			& +shuffle 50\%    & .9374 & .6221\\
			& +shuffle 75\%    & .8622 & .6574\\
			& +shuffle 100\%   & .9928 & .8521\\
			\hline
		\end{tabular}
	}\vspace{-.3cm}
	\caption{$R^2$ correlation obtained by fitting a linear regression of the encoding/decoding step with hidden states.}\label{tab:correlation}
\end{table}

\section{Results}\label{sec:results}

\vspace{-.2cm}
\noindent\textbf{Overall Performance.}  Table~\ref{tab:bleu} presents the BLEU scores of dialog machine translation systems. In open-domain dialog, BLEU-2 exhibits some (not large) correlation with human satisfaction, although BLEU scores are generally low.
For machine translation, we achieved 27.2 BLEU for the normal setting, which is comparable to 28.4 achieved by a baseline method in \newcite{baseline}. 

If we begin to shuffle the translation dataset, we see that BLEU drops gradually and finally reaches near zero if the training set is completely random (100\% shuffled). The results are not surprising and also reported in~\newcite{shuffle}. This provides a quick understanding on how the Seq2Seq is influenced by shuffled data.

\smallskip
\noindent\textbf{Length, Negative Log-Probability, and Entropy}. We now compare the length, probability, and entropy of dialog and translation systems, as well as the shuffling setting (Table~\ref{tab:stat}). The \textit{length} metric counts the number of words in a generated reply.\footnote{In some cases, an RNN fails to terminate by repeating a same word. Here, we assume a same word can be repeated at most four times.}
The \textit{negative log-probability} is computed as $-\frac1{|R|}\sum_{w\in R}\log p_\text{train}(w)$, where $R$ denotes all replies and $p_\text{train}(\cdot)$ is the unigram distribution of words in the training set. \textit{Entropy} is defined as $-\sum_{w\in R} p_\text{gen}(w)\log p_\text{gen}(w)$, where $p_\text{gen}(\cdot)$ is the unigram distribution in generated replies. Intuitively, both negative log-probability and entropy evaluate how much ``content'' is contained in the replies. These metrics are used in previous work~\cite{variational,seq2BF},\footnote{In our previous work~\cite{seq2BF}, the negative log-probability is mis-interpreted as entropy after email correspondence with some other peer researcher.} and obviously most relevant to our research question.

We first compare the dialog system with machine translation, both in a normal setting (no shuffling). We observe that, the dialog system does generate short and meaningless replies with lower length, negative log-probability, and entropy metrics than references, as opposed to machine translation where Seq2Seq's generated sentences are comparable to references in terms of these statistics. Quantitatively, the length is 20\% shorter than references. The negative log-probability and entropy decrease by 0.71 and 0.99, respectively; a decrease of 1 in negative log-probability and entropy metrics is large because they are logarithmic metrics. Although with a well-engineered Seq2Seq model (with attention, beam search, etc.), the phenomenon is less severe than a vanilla Seq2Seq in \newcite{seq2BF}, it is still perceivable and worth investigating.

We then applied the shuffling setting to the translation system. With the increase of shuffling rate, the Seq2Seq translation model precisely exhibits the phenomenon as a dialog system: the length decreases, the negative log-probability decreases, and the entropy decreases. In particular, the decreasing negative log probability implies that the generated words are more frequently appearing in the training set, whereas the decreasing entropy implies that the distribution of generated sentences spread less across the vocabulary.
In other words, artificially constructing an unaligned property in translation datasets---with all other settings remain unchanged---enables to reproduce the phenomenon in a dialog system. This shows evidence that the unaligned property could be one reason that causes the problem of short and meaningless replies in a dialog system.

\smallskip
\noindent\textbf{Correlation between Time Step and Hidden States.}
\newcite{MTlength} conduct an empirical study analyzing ``Why Neural Translations are the Right Length?'' They observe that, even the semantic of translation is not good, the length of generated reply is likely to be correct. They further find that some dimensions in RNN states are responsible for memorizing the current length in the process of sequence generation; the result is also reported in \newcite{visualizeRNN} previously. \newcite{MTlength} apply linear regression to predict the time step during sequence modeling based on hidden states, and compute the $R^2$ correlation as a quantitative measure.

Since a dialog system usually generates short replies (and thus not right length), we are curious what the $R^2$ correlation would be in a dialog system as well as shuffled translation settings. The results are shown in in Table~\ref{tab:correlation}. We find that the dialog system exhibits low correlation, and that the correlation also decreases in machine translation if data are shuffled (but not as worse as dialog systems). One inconsistent result, however, is that for the 100\% shuffled dataset, the correlation in the encoder side becomes 99\%, while the decoder correlation also increases to 85\%. We currently do not have good explanation to this.

\section{Conclusion and Discussion}\label{sec:conclusion}

In this paper, we addressed the question why dialog systems generate short and meaningless replies. We managed to reproduce this phenomenon in a well-behaving translation system by shuffling training data, artificially mimicking the scenario that a source sentence can have multiple equally plausible target sentences.

Admittedly, it is impossible to construct exactly the same scenario as dialog by using translation datasets (otherwise the translation just becomes dialog). However, the unaligned property is a salient difference, and by controlling this, we observe the desired phenomenon. Therefore it could be one cause of short and meaningless replies in dialog systems.

Our findings also explain why referring to additional information---including dialog context~\cite{context}, keywords~\cite{seq2BF} and knowledge bases~\cite{KB}---helps dialog systems: the number of plausible target sentences decreases if the generation is conditioned on more information; this intuition is helpful for future development of Seq2Seq dialog systems. Moreover, our experiments suggest that Seq2Seq models are more suitable to applications where the source and target information is aligned.

\section*{Acknowledgments}

We would like to thank Daqi Zheng and Yiping Song for helpful discussion.
\bibliography{naaclhlt2018}
\bibliographystyle{acl_natbib}

\appendix
\section{Experimental Setup}\label{appdx:setup}

\subsection{Neural Network}

We use the neural network in \newcite{toolkit} as our model. The encoder is a bidirectional recurrent neural network with gated recurrent units (GRUs). Let us consider one direction $\bm h_{t}=\GRU(\bm h_{t-1}, \bm x_t)$, where $\bm x_t$ is the input embedding at the time step $t$ and 
$\bm h_t$ is the hidden state. The computation of one step is given by
\begin{align}\nonumber
\bm z_t &= \sigma(W_z\bm x_t + U_z\bm h_{t-1})\\\nonumber
\bm r_t &= \sigma(W_r\bm x_t + U_r\bm h_{t-1})\\\nonumber
\widetilde{\bm h}_t &= \tanh(\bm r_t\circ U\bm h_{t}+W\bm x_t)\\\nonumber
\bm h_t &= (1-\bm z_t)\circ \bm \widetilde{\bm h}_t + \bm z_t\circ \bm h_{t-1}
\end{align}
where $W$'s and $U$'s are weights; $\sigma$ is the $\operatorname{sigmoid}$ function and $\circ$ is element-wise product.

Applying GRU-RNN to both directions and concatenating the resulting hidden states, we obtain the representation of the $i$th word in the source as
$\bm h_i^\text{(src)}=[\bm h_i^\text{($\overrightarrow{\text{src}})$}; \bm h_i^\text{($\overleftarrow{\text{src}})$}]$

The decoder is an RNN with two blocks of GRUs and an attention mechanism sandwiched in between. 
The first block of GRU computes an intermediate representation for the $j$th word in the target as
$\bm h_j'{}^\text{(tar)}=\GRU_1(\bm y_{j-1}, \bm h_j^\text{(tar)})$, where $\bm y_{j-1}$ is the embedding of the last word $y_{j-1}$. $\bm h_j'{}^\text{(tar)}$ is used to compute attention vector as

\begin{align}\nonumber
\widetilde{\alpha}_{ji}&=\bm v_a^\top \tanh(U_a\bm h_j'{}^\text{(tar)}+W_a \bm h_i^\text{(src)})\\\nonumber
\alpha_{ji}&=\frac{\exp\{\widetilde{\alpha}_{ji}\}}
	{\sum_k\exp\{\widetilde{\alpha}_{ki}\}}
\end{align}
A context vector is computed as
\begin{equation}
\nonumber
\bm c_j =\sum_i\alpha_{ji}\bm h_i^\text{(src)}
\end{equation}
Then $c_j$ is fed to the second block of GRU as
\begin{align}\nonumber
\bm h_j^\text{(tar)}=\GRU_2(\bm h_j'{}^\text{(tar)}, \bm c_j)
\end{align}

Finally, $\bm h_j$, $\bm y_{j-1}$, and $\bm c_j$ are fed to a fully connected layer and a softmax layer for prediction of the word $y_j$ at the time step $j$ in the decoder.

\subsection{Hyperparameter Settings}
In our all experiments, word embeddings were 512d. We used Adam to optimize all parameters, with initial
learning rate 0.0001. The dropout rate was set to 0.2. We set the mini-batch size to 60 to fit to GPU memory. 
In machine translation, RNN was 1024d and the vocabulary size was 30k in each language, whereas in the dialog model,
the RNN was 1000d and the vocabulary size was 50k.
For prediction beam search (beam size 12) was adopted to generate a translation or a reply.

\end{document}